\pdfoutput=1

\documentclass[11pt]{article}

\usepackage{emnlp2021}

\usepackage{times}
\usepackage{booktabs}
\usepackage{graphics}
\usepackage{graphicx}
\usepackage{latexsym}

\usepackage[T1]{fontenc}

\usepackage[utf8]{inputenc}

\usepackage{microtype}

\usepackage{amsmath,amsfonts,bm}

\def\eqref#1{equation~\ref{#1}}

\def\1{\bm{1}}

\def\vc{{\bm{c}}}

\def\ve{{\bm{e}}}

\def\mM{{\bm{M}}}

\def\mU{{\bm{U}}}
\def\mV{{\bm{V}}}

\DeclareMathAlphabet{\mathsfit}{\encodingdefault}{\sfdefault}{m}{sl}
\SetMathAlphabet{\mathsfit}{bold}{\encodingdefault}{\sfdefault}{bx}{n}

\newcommand{\R}{\mathbb{R}}

\usepackage{cleveref}
\Crefname{equation}{Eq.}{Eqns.}
\Crefname{figure}{Fig.}{Figs.}
\Crefname{theorem}{Thm.}{Thms.}

\title{A Simple and Effective Method To Eliminate the Self Language Bias in Multilingual Representations}

\author{Ziyi Yang$^1$\thanks{$\;\;$Work done during internship at Google Research.}$\;$, Yinfei Yang$^2$, Daniel Cer$^2$, Eric Darve$^1$ \\
$^1$Stanford University\\
\texttt{\{ziyi.yang,darve\}@stanford.edu}\\
$^2$Google Research\\
\texttt{\{yinfeiy,cer\}@google.com}}

\begin{document}
\maketitle
\begin{abstract}
Language agnostic and semantic-language information isolation is an emerging research direction for multilingual representations models. We explore this problem from a novel angle of geometric algebra and semantic space. A simple but highly effective method ``Language Information Removal (LIR)'' factors out language identity information from semantic related components in multilingual representations pre-trained on multi-monolingual data. A post-training and model-agnostic method, LIR only uses simple linear operations, e.g. matrix factorization and orthogonal projection. LIR reveals that for weak-alignment multilingual systems, the principal components of semantic spaces primarily encodes language identity information. We first evaluate the LIR on a cross-lingual question answer retrieval task~(LAReQA), which requires the strong alignment for the multilingual embedding space.
Experiment shows that LIR is highly effectively on this task, yielding almost 100\% relative improvement in MAP for weak-alignment models. We then evaluate the LIR on Amazon Reviews and XEVAL dataset, with the observation that removing language information is able to improve the cross-lingual transfer performance.  
\end{abstract}

\section{Introduction}

Recently, large-scale language modeling has expanded from English to the multilingual setting (i.a., \citet{bert,xlm,xlm-r}). Although these models are trained with language modeling objectives on monolingual data, i.e. without cross-lingual information, these multilingual systems exhibit impressive zero-shot cross-lingual ability \citep{xtreme}. These observations raise many questions and provide insight for multilingual representations learning. First, how is the language identity information and the semantic information expressed in the representation? Understanding their relations and underlying geometric structure is crucial for insights into designing more effective multilingual embedding systems. Second, how can we factor out the language identity information from the semantic components in representations? In many application, e.g. cross-lingual semantic retrieval, we wish to only keep the semantic information. Third, what is the geometric relation between different languages?
Efforts have been made to answer these questions, e.g. \citet{artetxe2020cross, chung2020rethinking, lauscher2020zero}. Such prior work has addressed the problem at training time. In this work, we systematically explore a post-training method that can be readily applied to existing multilingual models.

\begin{figure*}
  \centering
  \includegraphics[width=0.64\linewidth]{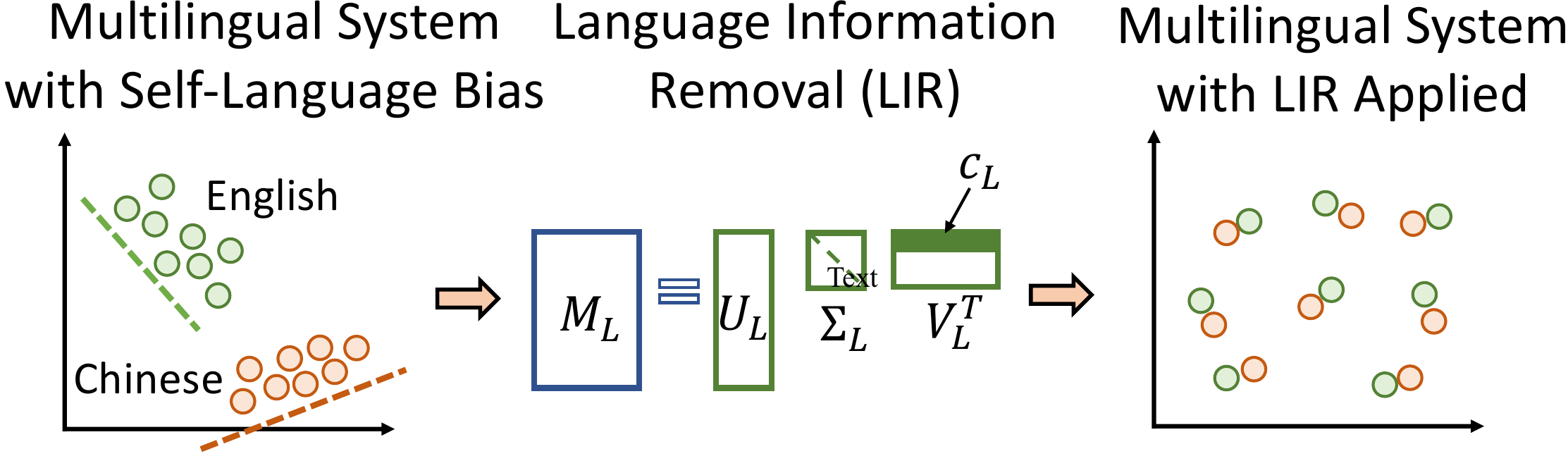}
  \vskip -0.1in
  \caption{Language Information Retrieval (LIR) removes language identification information using principle components of the original representation space. This mechanism is validated by LIR's effects demonstrated in \Cref{fig:pca}.}
  \label{fig:lir}
\vskip -0.2in
\end{figure*}

One of the first attempts in this research area, \citet{lareqa}, proposed two concepts for language agnostic models: weak alignment v.s. strong alignment. For multilingual systems with weak alignment, for any item in language $L_1$, the nearest neighbor in language $L_2$ is the most semantically ``relevant'' item. In the case of strong alignment, for any representation, all semantically relevant items are closer than all irrelevant items, regardless of their language. \citet{lareqa} show sentence representations from the same language tend to cluster in weak-alignment system. Similar phenomena can be observed on other pre-trained multilingual models like mBERT, XLM-R \citep{xlm-r} and CMLM \citep{cmlm}. \citet{lareqa} provides carefully-designed training strategies for retrieval-like model to mitigate this issue in order to obtain language agnostic multilingual systems. 

We systematically explore a simple post-training method we refer to as Language Information Removal (LIR), to effectively facilitate the language agnosticism in multilingual embedding systems.  
First introduced in \citet{cmlm} to reduce same language bias for retrieval tasks, the method uses only linear algebra factorization and post-training operation. LIR can be conveniently applied to any multilingual model. We show LIR yields surprisingly large improvements in several downstream tasks, including LAReQA, a cross-lingual QA retrieval dataset \citep{lareqa}; Amazon Reviews, a zero-shot cross lingual evaluation dataset; XEVAL, a collection of multilingual sentence embedding tasks. Our results suggest that the principal components of a multilingual system with self-language bias primarily encodes language identification information. Implementation for LIR is available at \url{https://github.com/ziyi-yang/LIR}.

\section{Language Information Removal for Self Language Bias Elimination}
In this section we describe Language Information Removal (LIR) to address the self language bias in multilingual embeddings~\cite{cmlm}. 
The first step is to extract the language identity information for each language space. Given a multilingual embedding system $E$, e.g. multilingual BERT, and a collection of multilingual texts $\{t_{L}^{i}\}$, where $t_{L}^{i}$ denotes the $i$th phrase in the collection for the language $L$. We construct a language matrix $\mM_{L} \in \R^{n\times d}$ for language $L$, where $n$ denotes the number of sentences in language $L$ and $d$ denotes the dimension of the representation. The row $i$ of $\mM_{L}$ is the representation of $t_{L}^{i}$ computed by $E$. 

Second, we extract language identification components for each language. One observation in multilingual systems is that representations from the same language tend to cluster together (w.r.t representations in other languages), even though these representations have different semantic meanings. This phenomenon is also known as ``weak alignment''~\citep{lareqa}. The mathematical explanation for this clustering phenomenon is that representations in the same language have shared vector space components. We propose that these shared components essentially represent the language identification information. Removing these language components should leave semantic-related information in the representations. 

To remove the shared components, or the language identification from the representations, we leverage singular value decomposition~(SVD) which identifies the principal directions of a space. We use SVD instead of PCA since SVD is more stable numerically (e.g. for Läuchli matrix). Specifically, the SVD of a language matrix is $\mM_{L} = \mU_{L} \bm{\Sigma}_L \mV_{L}^{T}$, where the columns of $\mV_L \in \R^{d \times d}$ are the right singular vectors of $\mM_{L}$. We take first $r$ columns of $\mV_L$ as the language identification components, denoted as $\vc_L \in \R^{d\times r}$. Different values of $r$ are explored in the next experiments section. Language identification components are removed as follows. Given a multilingual representation $\ve_{L}$ in language $L$, we subtract the projection of $\ve_{L}$ onto $\vc_L$ from $\ve_{L}$, i.e.
\begin{equation}
    \ve_{L} := \ve_{L} - \vc_L\frac{\vc_L^T \ve_{L}}{\|\ve_{L}\|_2}
\label{eq:lir}
\end{equation}

\section{Experiments}   
In the following experiments, sentences used for extracting principle components are sampled from Wiki-40B~\citep{wiki40b}. We use 10,000 sentences per language. We notice performance initially increases as more sentences are used but then is almost unchanged after $n > 10,000$. We tried different samplings of $\{t_{L}^{i}\}$ and text resources other than Wiki-40B, e.g., Tatoeba~\cite{laser}. The minimal differences in performance suggest language components are stable over different domains.

\subsection{Cross-lingual Answer Retrieval}
\label{sec:lareqa}
We first examine LIR on LAReQA, a cross-lingual answer retrieval dataset containing 11 languages \citep{lareqa}. LAReQA consists of two retrieval sub-datasets: XQuAD-R and MLQA-R. XQuAD-R is built by translating 240 paragraphs in the SQuAD v1.1 dev set into 10 languages and converting them to retrieval tasks following the procedure from ReQA \citep{reqa}. Similarly, MLQA-R is constructed by converting MLQA \citep{mlqa} to QA retrieval. In other words, each question in LAReQA has 11 relevant answers, one in each language. Two retrieval models with self language bias are presented in the LAReQA original paper, i.e. \textbf{``En-En''} and \textbf{``X-X''}. Specifically, the multilingual model \textbf{``En-En''} finetunes mBERT for QA retrieval on the 80,000 English QA pairs from the SQuAD v1.1 train set using a ranking loss. The model \textbf{``X-X''} trains on the translation (into 11 languages) of the SQuAD train set. In one training example, the question and answer are in the same language. Since given a question query, all positive examples are within-language, \textbf{``En-En''} and \textbf{``X-X''} exhibit strong self-language bias and weak-alignment property.

For evaluation, we first compute the language identification components with \textbf{``En-En''} and \textbf{``X-X''} models released by LAReQA. For testing, language identification components are removed from question and answer embeddings following \Cref{eq:lir}. Results are shown in \Cref{tab:lareqa} and the evaluation metric is mean average precision (MAP) of retrieval. Detailed results for each language are provided in the appendix (\Cref{tab:lareqa_apd}). Simply applying LIR results in significant improvements, almost $100\%$ relatively for \textbf{``X-X''} model on XQuAD-R. This huge boost reveals the algebraic structure for multilingual representation space: in weak-alignment multilingual system, the principal components primarily encode language information. In LAReQA, each language has one of the relevant answers. The performance improvement itself already indicates less language bias. 

\begin{table}[htb]
\centering
\small
\vskip -.1in
\begin{tabular}[c]{l|cc|cc}
\toprule
 & \multicolumn{2}{c|}{XQuAD-R} & \multicolumn{2}{c}{MLQA-R} \\
        & \textbf{En-En} & \textbf{X-X}   & \textbf{En-En} & \textbf{X-X} \\ \midrule
w/o LIR  & 27.8 & 23.3 & 35.7 & 26.0 \\ 
$r = 1$ & \textbf{36.7} & 45.2 & \textbf{37.0} & \textbf{42.4} \\
$r = 2$ & 36.7 & 45.6 & 36.2 & 41.6 \\
$r = 3$ & 36.5 & \textbf{45.9} & 36.3 & 41.6 \\
$r = 4$ & 36.4 & 45.7 & 36.1 & 41.4 \\
\bottomrule
\end{tabular}
\vskip -.1in
\caption{Mean average precision (MAP) of model \textbf{``En-En''} and \textbf{``X-X''} with and without LIR.}
\label{tab:lareqa}
\vskip -.1in
\end{table}

\begin{figure}
  \centering
  \includegraphics[width=\columnwidth]{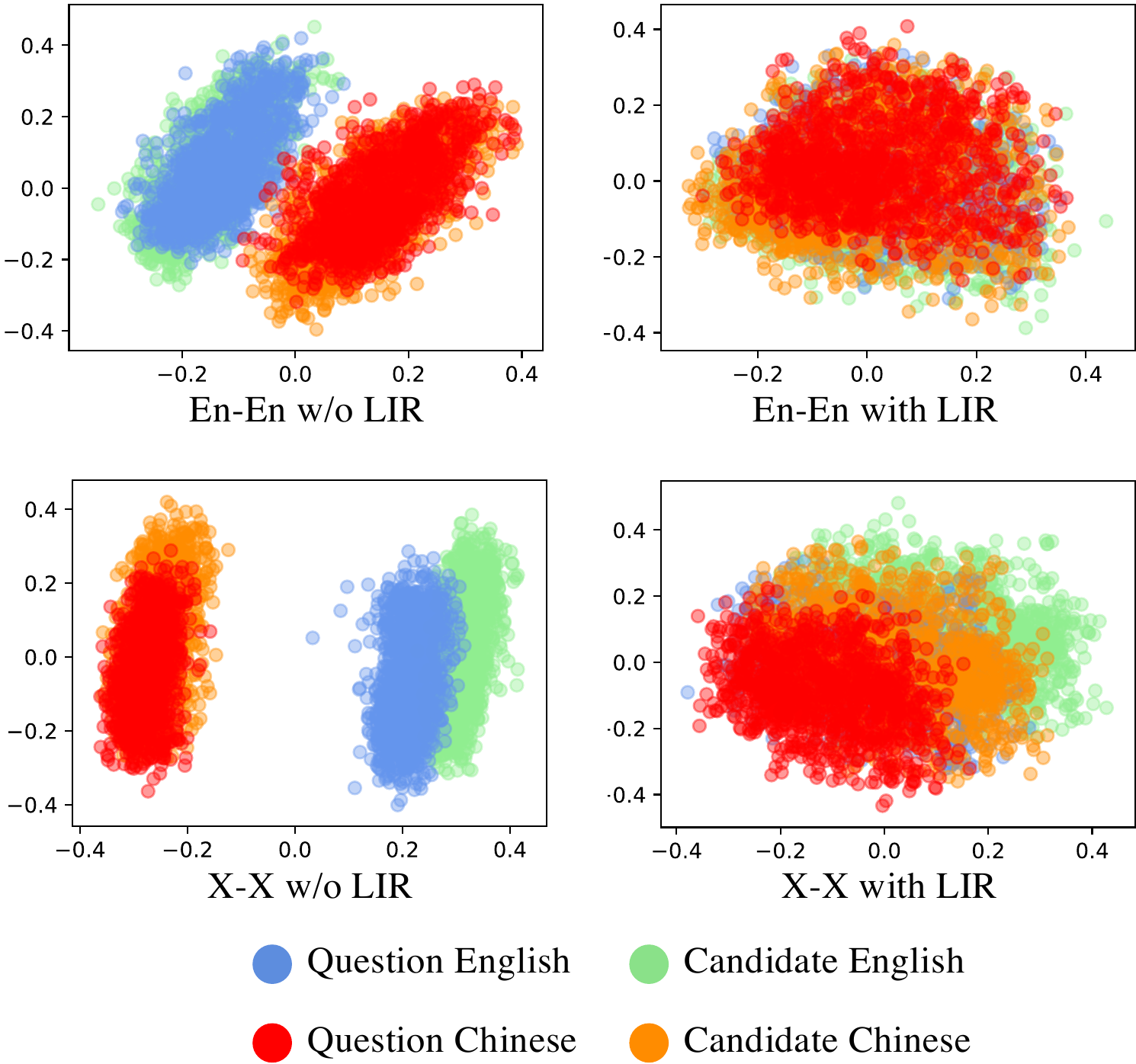}
  \vskip -0.1in
  \caption{PCA projections of English and Chinese embeddings on the XQuAD-R dataset, with and without LIR. The two subfigures on the left are reproduced by authors to follow Figure 5 in \citet{lareqa}.}
  \label{fig:pca}
\vskip -0.2in
\end{figure}

To further illustrate the effect of LIR, we plot the 2D PCA projection of questions and candidates in Chinese and English for the XQuAD-R dataset. Without LIR, as plotted on the left of \Cref{fig:pca}, Chinese and English embeddings are separated while questions and candidates in the same language are clustering together\footnote{Note these two subfigures on the left are reproduced by authors to imitate Figure 5 in \citet{lareqa} in order to better demonstrate the effects of LIR}. This weak-alignment property is especially prominent for model \textbf{``X-X''}. After applying LIR, the separation between the two languages vanishes. Questions and candidates embeddings, no matter which language they are in, group together. Both model \textbf{``En-En''} and \textbf{``X-X''} now exhibit strong cross-lingual alignment.

\subsection{Amazon Reviews}
\label{sec:amazon}
We further evaluate LIR on zero-shot transfer learning with Amazon Reviews Dataset \citep{amazon}. In this subsection, we use multilingual BERT \citep{bert} as the embedding model. Following \citet{chidambaram}, the original dataset is converted to a classification benchmark by treating reviews of more than 3 stars as positive and negative otherwise. We split 6000 English reviews in the original training set into 90\% for training and 10\% for development. A logistic classifier is trained on the English training set and then evaluated on English, French, German and Japanese test sets (each has 6000 examples) using the same trained model, i.e. the evaluation is zero-shot.
The weights for mBERT are fixed. The representation of a sentence/phrase is computed as the average pooling of the transformer encoder outputs. LIR is applied in both training and evaluation stage using the corresponding language components.

Results presented in \Cref{tab:amazon} show that removing the language components from multilingual representation is beneficial for cross-lingual zero-shot transfer learning of mBERT. LIR is expected to leave only semantic-related information in the representation so that the logistic classifier trained on English should be conveniently transferred to other languages. Another interesting observation is that unlike semantic retrieval, the peak performance usually occurs at $r > 1$.

\begin{table}[htb]
\centering
\small
\begin{tabular}[c]{l|ccccc}
\toprule
      & en & de & fr & jp & Avg. \\ \midrule
w/o LIR  & 80.0 & 70.4 & 73.1 & 71.7 & 73.8 \\ 
$r = 1$  & 80.2 & 70.9 & 74.6 & 72.6 & 74.6\\
$r = 2$  & \textbf{80.5} & 70.9 & 75.6 & \textbf{73.1} & \textbf{75.0} \\
$r = 3$  &  80.2 & 70.8 & 75.4 & 71.8 & 74.5 \\
$r = 5$  &  80.2 & 70.8 & \textbf{76.1} & 72.0 & 74.8 \\
$r = 8$  &  80.2 & \textbf{71.1} & 75.2 & 70.8 & 74.3 \\
$r = 10$ &  80.3 & 71.0 & 76.0 & 71.2 & 74.6 \\
$r = 12$ &  80.0 & 70.9 & 76.0 & 71.4 & 74.6 \\
\bottomrule
\end{tabular}
\vskip -.1in
\caption{Classification accuracy on Amazon Reviews Dataset.}
\label{tab:amazon}
\vskip -0.2in
\end{table}


\subsection{XEVAL}
\label{sec:xeval}
We have tested LIR on cross-lingual benchmarks in previous sections. In this section, we apply LIR in XEVAL, a collection of multilingual sentence representation benchmark \citep{cmlm}. The training set and test set of XEVAL are in the \textbf{same} language (i.e. the evaluation is not cross-lingual). Benchmarks on XEVAL include Movie Reviews \citep{mr}, binary SST (sentiment analysis, \citet{sst}), MPQA (opinion-polarity, \citet{mpqa}), TREC (question-type, \citet{trec}), CR (product reviews, \citet{cr}), SUBJ (subjectivity/objectivity, \citet{subj}) and SICK (both entailment and relatedness \citep{sick}). For this evaluation, we use mBERT as the base multilingual encoder. Still the weights of mBERT are fixed during training and only downstream neural structures are trained. The training, cross-validation and evaluation uses SentEval toolkit \citep{senteval}. 

Results are presented in \Cref{tab:xeval}. The metric is the averaging performance across 9 datasets mentioned above. Introducing LIR is beneficial on German, Spanish, French and Chinese. We also notice that for English dataset, removing principal components actually hurts the performance. This observation also echoes with findings in previous English sentence embedding works, e.g. \citet{gem}. We speculate this is because English data are dominant in mBERT training data. Therefore mBERT representations exhibit similar behaviors with monolingual English sentence embeddings.

\begin{table}[htb]
\centering
\small
\vskip -.12in
\resizebox{\columnwidth}{!}{
\begin{tabular}[c]{l|cccccc}
\toprule
         & en & de & es & fr & zh & Avg. \\ \midrule
w/o LIR  & \textbf{80.8} & 78.1 & 78.8 & 79.1 & 79.3 & 79.2\\ 
$r = 1$  &  80.4 & 78.2 & 79.0 & 79.1 & 79.3 & 79.2 \\
$r = 2$  &  80.7 & \textbf{78.5} & \textbf{79.4} & \textbf{79.3} & \textbf{79.4} & \textbf{79.5}\\
$r = 5$  &  80.6 & 78.0 & 79.4 & 78.9 & 79.3 & 79.2 \\
$r = 10$ &  80.2 & 78.4 & 79.0 & 79.0 & 78.9 & 79.1 \\
\bottomrule
\end{tabular}
}
\vskip -.1in
\caption{Results of applying LIR to XEVAL dataset. The metric is the average of 9 downstream tasks.}
\label{tab:xeval}
\vskip -.2in
\end{table}

\subsection{Application to Models without Self-Language Bias}
In previous sections, we have shown the great effectiveness of LIR on weak-alignment systems. As an additional analysis, we examine LIR on multilingual models without self language bias, i.e. models \textbf{``X-X-mono''} and \textbf{``X-Y''} introduced in the original LAReQA paper. Model \textbf{``X-X-mono''} is modified from \textbf{``X-X''} by ensuring that each training batch is monolingual so that in-batch negative and positive examples are in the same language. In model \textbf{``X-Y''}, questions and answers are allowed to be translated to different languages, which directly encourage the model to regard answers in different languages from the question as correct. With such designs in training, \textbf{``X-X-mono''} and \textbf{``X-Y''} are shown to be without self-language bias, i.e. semantically relevant representations are closer than all irrelevant items, regardless of their languages.

The evaluation process is similar as in \Cref{sec:lareqa}. Results are presented in \Cref{tab:wo_bias}. Applying LIR leads to a slight performance decrease for X-X-mono. While the drop in X-Y is notable and we suspect this is because the training process for X-Y avoids, by design, self-language bias. Rather, the principal components of X-Y contain essential semantic-related information for the retrieval task. This result is \textbf{not} negative and actually support our argument, since for ``strong alignment'' multilingual systems, principal components should both contain semantic and language-related information. Then removing principal components will hinder the semantic retrieval. For weak-alignment models, removing just the first component should be adequate for cross-lingual retrieval (\cref{tab:lareqa}). For tasks like classification and sentiment analysis (\cref{tab:amazon,tab:xeval}), the optimal number of components to remove seems to vary on different datasets.

\begin{table}[htb]
\centering
\small
\begin{tabular}[c]{l|cc|cc}
\toprule
 & \multicolumn{2}{c|}{XQuAD-R} & \multicolumn{2}{c}{MLQA-R} \\
        & \textbf{X-X-mono} & \textbf{X-Y}   & \textbf{X-X-mono} & \textbf{X-Y} \\ \midrule
w/o LIR  & \textbf{50.8} & \textbf{62.6} & 48.6 & \textbf{48.5} \\ 
$r = 1$ & 50.6 & 59.5 & \textbf{48.8} & 46.2 \\
$r = 2$ & 49.8 & 58.1 & 48.0 & 45.5 \\
$r = 3$ & 49.3 & 57.1 & 47.8 & 44.8 \\
$r = 4$ & 48.9 & 56.5 & 47.4 & 44.2 \\
\bottomrule
\end{tabular}
\vskip -.1in
\caption{Mean average precision (MAP) of \textbf{``X-X-mono''} and \textbf{``X-Y''} models without language bias.}
\label{tab:wo_bias}
\vskip -.2in
\end{table}

\section{Related Work \& Our Novelty}
Different training methods have been proposed to obtain language agnostic representations. LASER \citep{laser} leverages translation pairs and BiLSTM encoder for multilingual sentence representation learning. Multilingual USE \citep{muse} uses training data such as translated SNLI, mined multilingual QA and translation pairs to learn multilingual sentence encoder. AMBER \citep{amber} aligns contextualized representations of multilingual encoders at different granularities. LaBSE \citep{labse} finetunes a pretrained language model with the bitext retrieval task and mined cross-lingual parallel data to obtain language agnostic sentence representations. In contrast, LIR does not require any parallel data for semantic alignment.

\citet{cca} propose a canonical correlation analysis (CCA) based method to add multilingual context to monolingual embeddings. The method is post-processing and requires bilingual word translation pairs to determine the projection vectors. In contrast, LIR is post-training and does not require labeled data. \citet{mrkvsic2017semantic} build semantically specialized cross-lingual vector spaces. Like CCA, their methods requires the additional training to adjust the original embeddings using supervised data: cross-lingual synonyms and antonyms. \citet{libovicky2019language} propose that the language-specific information of mBERT is the centroid of each language space (the mean of embeddings). \citet{zhao-etal-2021-inducing} propose several training techniques to obtain language-agnostic representations, including segmenting orthographic tokens in training data and aligning monolingual spaces by training. In contrast, LIR is post-training and model-agnostic. Critically, this means LIR can be conveniently and easily applied to any trained multilingual systems without further training. 

Previous explorations on principal components of the semantic space for sentence embeddings include \citet{sif} and \citet{gem}, whereby principal component removal is investigated for monolingual models and the evaluation is only conducted on semantic similarity benchmarks. In contrast, our work investigates the multilingual case and the evaluation is more diverse, e.g. cross-lingual transfer learning. \citet{mu2018all} explore removing top components from English representations. However, it was unclear prior to our work what purpose is served by removing principal components within multilingual and cross-lingual settings. We demonstrate these principal components represent language information for weak-alignment multilingual models.

Compared with \citet{cmlm}, the novelty of this work is two-fold. First, it is unclear in \citet{cmlm} whether the assumption (i.e. principal components contain language information) holds true for both weak and strong-alignment multilingual models. In this work we clearly show that it is is valid for weak-alignment models (\Cref{sec:lareqa}). However, for strong-alignment systems, the assumption is not quite true (\Cref{tab:wo_bias}). Second, in \citet{cmlm}, the evaluation is only conducted on Tatoeba, a semantic retrieval dataset. While in this work, evaluations are more comprehensive. Besides the cross-lingual retrieval dataset LAReQA, our experiments include cross-lingual zero-shot learning (\Cref{sec:amazon}) and monolingual transfer learning (\Cref{sec:xeval}). These extra results establish the effectiveness of LIR beyond the domain of semantic retrieval.

\section{Conclusion}
In this paper, we investigate the self-language bias in multilingual systems. We explore a simple method ``Language Identity Removal (LIR)''. This method identifies and removes the language information in multilingual semantic space by singular value decomposition and orthogonal projection. Although as a simple and linear-algebra-only method, LIR is highly effective in several downstream tasks, including zero-shot transfer learning, sentiment analysis, etc. Especially for cross-lingual retrieval, introducing LIR increases the performance of weak-alignment multilingual systems by almost 100\% relatively in MAP.

\section*{Acknowledgments}
We would like to thank anonymous reviewers for their comments, as well as our teammates from Descartes, Google Brain and other Google teams for their valuable feedback.

\bibliography{anthology,custom}
\bibliographystyle{acl_natbib}

\clearpage
\appendix

\section{Experimental results for each language of model \textbf{``X-X''} on LAReQA}
Here we provide the detailed experiment results of each language on the XQuAD-R dataset. The multilingual encoder is model \textbf{``X-X''}.
\begin{table}[htb]
\centering
\small
\begin{tabular}[c]{l|ccccc}
\toprule
      & w/o LIR & $r = 1$ & $r = 2$ & $r = 3$ & $r = 4$ \\ \midrule
 ar & 20.5 & \textbf{40.5} & 40.4 & 40.4 & 40.0 \\ 
 de & 27.5 & 48.3 & \textbf{49.8} & 49.7 & 49.6 \\
 el & 20.9 & 43.5 & 43.9 & 44.1 & \textbf{44.2} \\
 en & 27.3 & 55.1 & 55.0 & \textbf{55.3} & \textbf{55.3} \\
 es & 27.6 & 52.6 & \textbf{52.8} & 52.7 & 52.6 \\
 hi & 18.6 & 36.5 & 36.8 & \textbf{37.5} & 37.3 \\
 ru & 24.9 & \textbf{48.2} & 49.6 & 49.6 & 49.4 \\
 th & 16.8 & 34.7 & \textbf{35.1} & 34.9 & 34.6 \\
 tr & 23.8 & 45.3 & 45.4 & \textbf{46.3} & 46.2 \\
 vi & 24.8 & \textbf{49.2} & 48.9 & 48.9 & 48.6 \\
 zh & 24.7 & 43.8 & 43.8 & \textbf{45.3} & 45.2 \\
\bottomrule
\end{tabular}
\caption{Experimental results for each language of model \textbf{``X-X''} on the XQuAD-R dataset.}
\label{tab:lareqa_apd}
\end{table}



\end{document}